\documentclass[runningheads]{llncs}
\usepackage{graphicx}

\graphicspath{{figures/}}
\usepackage{comment}
\usepackage{epsfig}
\usepackage{amsmath}
\usepackage{amssymb}
\usepackage{array}
\usepackage{tabu}
\usepackage{colortbl}
\definecolor{mygray}{gray}{.9}
\begin{document}
\title{Factored Attention and Embedding for Unstructured-view Topic-related Ultrasound Report Generation}
%
%
\author{Fuhai Chen$^1$, Rongrong Ji$^{12}$\thanks{corresponding author}, Chengpeng Dai$^{1}$, Xuri Ge$^{1}$,\\ Shengchuang Zhang$^{1}$, Xiaojing Ma$^{3}$, Yue Gao$^{4}$\\$^1$School of Informatics, Xiamen University, $^2$Peng Cheng Laboratory,\\ $^3$Wuhan Asia Heart Hospital, $^4$School of Software, Tsinghua University}
\institute{\email{\{cfh3c,daiccc,gexuri\}@stu.xmu.edu.cn, \{rrji,zsc\_2016\}@xmu.edu.cn, gaoyue@tsinghua.edu.cn}}

%
\maketitle              
\begin{abstract}
Echocardiography is widely used to clinical practice for diagnosis and treatment, \emph{e.g.}, on the common congenital heart defects. The traditional manual manipulation is error-prone due to the staff shortage, excess workload, and less experience, leading to the urgent requirement of an automated computer-aided reporting system to lighten the workload of ultrasonologists considerably and assist them in decision making. Despite some recent successful attempts in automatical medical report generation, they are trapped in the ultrasound report generation, which involves unstructured-view images and topic-related descriptions. To this end, we investigate the task of the unstructured-view topic-related ultrasound report generation, and propose a novel factored attention and embedding model (termed \emph{FAE-Gen}). The proposed FAE-Gen mainly consists of two modules, \emph{i.e.}, view-guided factored attention and topic-oriented factored embedding, which 1) capture the homogeneous and heterogeneous morphological characteristic across different views, and 2) generate the descriptions with different syntactic patterns and different emphatic contents for different topics. Experimental evaluations are conducted on a to-be-released large-scale clinical cardiovascular ultrasound dataset (CardUltData). Both quantitative comparisons and qualitative analysis demonstrate the effectiveness and the superiority of FAE-Gen over seven commonly-used metrics.
\end{abstract}

\section{Introduction}

Echocardiography is widely used in hospitals for the diagnosis of common congenital heart defects in both children and adults, such as ventricular septal defect (VSD) \cite{minette2006ventricular} and atrial septal defect (ASD) \cite{webb2006atrial}. An ultrasonologist completes an ultrasonic diagnostic report ahead of ultrasound scanner by analyzing the ultrasound images in different sections (views), where the images of each view record the different blood flow movements. However, this process of medical image interpretation and reporting can be error-prone due to staff shortage, excess workload, and less experience \cite{seward2002hand,niendorff2005rapid,lucas2009diagnostic,stokke2014brief}. Therefore, an automated computer-aided reporting system is urgently required to reduce workload and error occurrences, where the ultrasound images are taken as inputs and the diagnostic report is automatically generated.

\begin{figure}[t]
\centering
\includegraphics[width=1.0\linewidth]{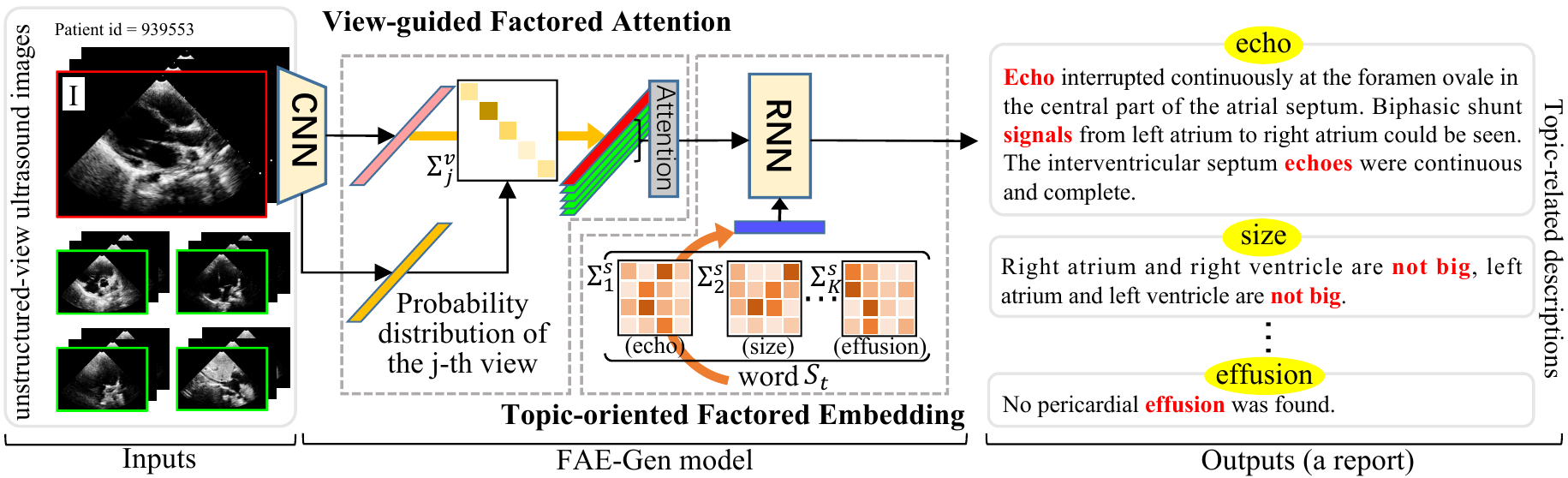}
\setlength{\abovecaptionskip}{-12pt}
\caption{The overview of the proposed \emph{FAE-Gen} for automatical unstructured-view topic-related ultrasound report generation. \label{fig:framework}}
\vspace{-4mm}
\end{figure}

Recently, there are increasing works attracted in the medical report generation problem \cite{zhang2017mdnet,xue2018multimodal,li2018hybrid,jing2018automatic}. Typically, inspired by image captioning \cite{vinyals2015show}, they achieve the system on the pairs of the medical images and the corresponding reports via a deep learning technology based encoder-decoder architecture, where the features of a medical image are first extracted by a convolutional neural network (CNN) \cite{krizhevsky2012imagenet,he2016deep} and then fed into recurrent neural networks (RNNs) \cite{hochreiter1997long} to generate the word sequences as a diagnostic report. For example, Jing \emph{et al.} \cite{jing2018automatic} proposed a co-attention model, where the visual (morphological) features and the tag features are extracted from CNN and jointly attended into hierarchical RNNs \cite{krause2017hierarchical} to describe the chest radiology images. Most of these works focused on generating a generally-described report upon one or two structured-view (frontal or lateral) images, like radiology report generation task \cite{xue2018multimodal,jing2018automatic}, which, however, can not handle the case when the views of the medical images are unstructured\footnote{\footnotesize \textbf{Defined as}: 1) the identifications of views are unknowable, 2) some views are missing, and 3) the ultrasound images (video frames) of each view vary a lot due to blood flow movements).} and the descriptions are topic-related in ultrasound report generation as shown in Fig. \ref{fig:framework}.



In the automatical ultrasound report generation, there are two key issues for the unstructured-view inputs and the topic-related outputs. On the one hand, the ultrasound images of different views have their homogeneous morphological characteristic on the same pathology as well as their heterogeneous morphological characteristic over different views. Thus the morphological relevance and diversity of the across-view ultrasound images should be captured during the visual feature representation for the accurate and full-sided text generation. On the other hand, different topic-related diagnostic descriptions have different syntactic patterns and different emphatic contents. To this end, different topic related word embedding should be imported into RNN to generate the discriminative descriptions.

Driven by the above insights, we propose a novel factored attention and embedding model for automatical unstructured-view topic-related ultrasound report generation (termed \emph{FAE-Gen}) as detailedly illustrated in Fig. \ref{fig:framework}. To capture the morphological relevance and diversity of the across-view ultrasound images, we develop a view-guided factored attention module to disentangle the view factors from the chaotic feature transformation. To import the topic-related word embedding into discriminative diagnostic descriptions, we design a topic-oriented factored embedding module to decompose the word embedding matrix into a topic-specific matrix and two shared transformation matrices. Specially, given unstructured-view ultrasound images of a patient, for each ultrasound image, we first extract its morphological feature (in pink) and the corresponding probability distribution of the recognized view (in yellow) from a pre-trained CNN model. Then, in the view-guided factored attention module, the view distribution is diagonalized as a view-specific factor matrix, and its morphological feature is transformed into a view-related feature by composing of the factor matrix and the shared parameterized matrices. The new features of all images are attended into the state of RNN (Sec. \ref{ViewFA}). Meanwhile, in each state of RNN, the word vector is transformed into a topic-related word feature by composing of the topic-related parameterized factor matrix and the other two shared parameterized matrices in topic-oriented factored embedding module. (Sec. \ref{TopicFE}). In this way, the topic-related word features are fed into the states of RNN to generate a topic-related description. Finally, the overall model is trained upon the supervision of the real report descriptions in a end-to-end manner.

The contributions of our work are: 1) we are the first to deal with a new medical report generation problem, \emph{i.e.}, automatical ultrasound report generation, with unstructured-view image inputs and topic-related description outputs. 2) we propose a novel factored attention and embedding model (FAE-Gen) to capture the homogeneous and heterogeneous morphological characteristic across different views in the view-guided factored attention module, as well as to generate the topic-related diagnostic descriptions in the topic-oriented factored embedding module. 3) Our dataset with ultrasound images and reports will be released publicly to promote the research.

\section{Methodology}

The proposed FAE-Gen is based on the fundamental encoder-decoder architecture of image captioning \cite{vinyals2015show}, where an image $I$ is first encoded into a visual feature ${\rm\mathbf{v}}$ by utilizing the convolution neural network (CNN) and then the visual feature ${\rm\mathbf{h}}^{v}$ is fed into a recurrent neural network (RNN) to generate the sequential words $\hat{S}_{1:T}$ ($T$ denotes the textual length of the description). In the unstructured-view topic-related ultrasound report generation, suppose there are $M$ ultrasound images $I_{1:M}$ in different views and $K$ real descriptions $S^{1:K}$ with different topics, the objective of FAE-Gen is to maximize the log likelihood as below:

\vspace{-6mm}
\begin{align}
\log P(S^{k}|I_{1:M};\theta) = \sum_{t=0}^{T} \log p(S_t^{k}|{\rm\mathbf{h}}_{1:M}^{v}, {\rm\mathbf{y}}_{1:M},S_{0:t-1}^{k}),
\label{objective}
\end{align}
\noindent
where $k\in\{1,\ldots,K\}$ denotes one of the topics. $\theta$ denotes the model parameter set that needs to be learned during model training, \emph{i.e.}, the optimal $\theta^*$ can be obtained by $\arg\max_{\theta} \sum_{(I_{1:M},S^{k})} \log P(\hat{S}^{k} = S^{k}|I_{1:M};\theta)$. To simplify, we omit $\theta$ in the right side. ${\rm\mathbf{h}}_{1:M}^{v}$ and ${\rm\mathbf{y}}_{1:M}$ denote the morphological features and the probability distributions of different views, respectively. To extract ${\rm\mathbf{h}}_{1:M}^{v}$ and ${\rm\mathbf{y}}_{1:M}$, we modify a CNN model, ResNet-50 \cite{he2016deep}, with two softmax outputs corresponding to the category spaces of the cardiovascular diseases (ASD, VSD, and normal) and five views (see Sec. \ref{exp} for details), respectively. We pre-train the ResNet-50 under the supervision of the disease and view categories, after which ${\rm\mathbf{h}}_{1:M}^{v}$ can be obtained from the final fully-connected layer and ${\rm\mathbf{y}}_{1:M}$ can be predicted from the corresponding softmax output. We transform ${\rm\mathbf{h}}^{v}$ into a view-related feature by using ${\rm\mathbf{y}}$ (Sec. \ref{ViewFA}). To obtain the $t$-th word $\hat{S}_t$, we use the Bi-directional Long Short-Term Memory (Bi-LSTM)\footnote{\footnotesize Bi-LSTM is an advanced version of RNN, which is widely used in long sentences or paragraphes to better capture the textual contexts. In this paper, we still call it \emph{RNN} for readability.} \cite{graves2005framewise} to generate $S_t$ depending on ${\rm\mathbf{h}}_{1:M}^{v}$, ${\rm\mathbf{y}}_{1:M}$ and the preceding words $S_{0:t-1}$ ($S_0$ is a start sign) according to the chain rule. Commonly, $S_t$ is represented with a one-hot vector ${\rm\mathbf{x}}_t$ according to its index in the dictionary. The word feature ${\rm\mathbf{h}}_{t}^{s}$ is then obtained via ${\rm\mathbf{h}}_{t}^{s} = {\rm\mathbf{E}}{\rm\mathbf{x}}_t$, where ${\rm\mathbf{E}}$ is a embedding matrix \cite{mikolov2013efficient}. In our paper, ${\rm\mathbf{E}}$ decomposes into a topic-related factor matrix and other embedding matrices for topic-guided description generations (Sec. \ref{TopicFE}).


\subsection{View-guided Factored Attention\label{ViewFA}}

In the proposed FAE-Gen, there are multiple ultrasound images of different views as the inputs, leading to two key technical points. First, how to transform the morphological feature ${\rm\mathbf{h}}^{v}$ into view-related features for the enhanced and discriminative representation. Second, how to weight the importance of these view-specific features. To this end, we design a view-guided factored attention module as shown in Fig. \ref{fig:framework}, where the traditional transformation $\hat{\rm\mathbf{h}}^{v}={\rm\mathbf{W}}{\rm\mathbf{h}}^{v}$ is advanced as follows:

\vspace{-4mm}
\begin{align}
\hat{\rm\mathbf{h}}_j^{v} = {\rm\mathbf{U}}{\rm\mathbf{\Sigma}}_{j}^{v}{\rm\mathbf{V}}{\rm\mathbf{h}}_j^{v},
\end{align}
\noindent
where $j$ denotes the $j$-th view. ${\rm\mathbf{U}}$ and ${\rm\mathbf{V}}$ are the shared parameterized matrices among all the samples. ${\rm\mathbf{\Sigma}}_{j}^{v}$ is a view-related factor matrix obtained by $diag({\rm\mathbf{y}}_{j})$, which plays a role of view guidance for the transformation. To weight these view-specific visual features and feed them into the RNN, we calculate their relevance to the hidden feature in each state of RNN via a attention mechanism. For the $t$-th RNN state, formulated as:

\vspace{-4mm}
\begin{align}
a_{j,t} = {\rm\mathbf{W}}^{a}\sigma({\rm\mathbf{W}}^{v}\hat{\rm\mathbf{h}}_j^{v} + {\rm\mathbf{W}}^{z}{\rm\mathbf{h}}_{t-1}^{s}), \
{\rm\boldsymbol{\alpha}}_t = softmax({\rm\mathbf{a}}_t), \
{\rm\mathbf{h}}_{t}^{a} = \sum_{j=1}^{M}\alpha_{j,t}\hat{\rm\mathbf{h}}_j^{v},
\label{visualattention}
\end{align}
\noindent
where ${\rm\mathbf{W}}^{a}$, ${\rm\mathbf{W}}^{v}$, and ${\rm\mathbf{W}}^{z}$ are the shared parameter matrices of linear transformation. ${\rm\mathbf{h}}_{t-1}^{s}$ denotes the preceding hidden feature in the $t$-th RNN state (detailed in Sec. \ref{TopicFE}). $\sigma$ is non-linear function (we use hyperbolic tangent). ${\rm\boldsymbol{\alpha}}_t$ is the relevance vector. Finally, a weight-sum attention feature ${\rm\mathbf{h}}_{t}^{a}$ is obtained as Eq. \ref{visualattention}.

\subsection{Topic-oriented Factored Embedding\label{TopicFE}}

The traditional word embedding ${\rm\mathbf{h}}_{t}^{s} = {\rm\mathbf{E}}{\rm\mathbf{x}}_t|_{t=0}^{T}$ causes the single syntactic pattern and emphatic content due to its uniform embedding matrix ${\rm\mathbf{E}}$ for the word representation. To generate the topic-related description, we design a topic-oriented factored embedding module as shown in Fig. \ref{fig:framework}. Specifically, we use the Bi-LSTM as our RNN model. The formulations in the $t$-th RNN state are:

\vspace{-4mm}
\begin{equation}
\!\resizebox{0.93\linewidth}{!}{$\overrightarrow{\rm\mathbf{h}}_{t}^{s} = f(\overrightarrow{\rm\mathbf{A}}\overrightarrow{\rm\mathbf{\Sigma}}_k^{s}\overrightarrow{\rm\mathbf{B}}{\rm\mathbf{x}}_{t-1},\overrightarrow{\rm\mathbf{W}}^{s}{\rm\mathbf{h}}_{t}^{a}), \overleftarrow{\rm\mathbf{h}}_{t}^{s} = f(\overleftarrow{\rm\mathbf{A}}\overleftarrow{\rm\mathbf{\Sigma}}_k^{s}\overleftarrow{\rm\mathbf{B}}{\rm\mathbf{x}}_{t-1},\overleftarrow{\rm\mathbf{W}}^{s}{\rm\mathbf{h}}_{t}^{a}), {\rm\mathbf{h}}_{t}^{s} = g([\overrightarrow{\rm\mathbf{h}}_{t}^{s};\overleftarrow{\rm\mathbf{h}}_{t}^{s}])$},
\end{equation}
\vspace{-5mm}
\begin{equation}
S_t \thicksim {\rm\mathbf{p}}_t = softmax({\rm\mathbf{h}}_{t}^{s}),
\label{wordembedding}
\end{equation}
\noindent
where $\rightarrow$ and $\leftarrow$ denote the forward and the backward directions. $f$ is a generic function in LSTM, which includes input, forget, output, and cell-related functions \cite{hochreiter1997long}. $\overrightarrow{\rm\mathbf{A}}$, $\overrightarrow{\rm\mathbf{B}}$, $\overleftarrow{\rm\mathbf{A}}$, and $\overleftarrow{\rm\mathbf{B}}$ are the shared parameter matrices. $\overrightarrow{\rm\mathbf{\Sigma}}_k^{s}$ and $\overleftarrow{\rm\mathbf{\Sigma}}_k^{s}$ are the $k$-th topic related parameter matrices, which are optimized with the topic-related descriptions. All the above matrices are initialized randomly. Bidirectional $\overrightarrow{\rm\mathbf{h}}_{t}^{s}$ and $\overleftarrow{\rm\mathbf{h}}_{t}^{s}$ are then mapped into ${\rm\mathbf{h}}_{t}^{s}$ via a non-linear function $g$. Finally, the current word $S_t$ can be sampled according to the word probability distribution ${\rm\mathbf{p}}_t$.


\section{Experiments\label{exp}}

In this section, we first introduce the proposed dataset on cardiovascular ultrasound images. Then we describe the experimental settings. Finally, we evaluate and discuss the performance of the proposed method on the cardiovascular ultrasound report generation task.

\paragraph{\textbf{Dataset.}} We obtain the original cardiovascular ultrasound images from the specialized hospital. The patients' heart dynamic images are recorded through the Philips ultrasound machine. Comprehensive evaluation of the cardiac structure and function in patients by two dimensional echocardiography, color Doppler echocardiography and M-mode echocardiography. And Images and data are stored in Philips IE33 and 7C ultrasound machines. Images and data are diagnosed by an experienced ultrasound physician who has worked for more than 2 years analyzing images and data. All the private information of the patients have been processed. For cardiovascular diseases, doctors typically diagnose patients with observed data of echocardiography, and obtain three types of diagnostic results: normal, VSD, ASD. VSD refers to ventricular septal defect, and ASD refers to atrial septal defect. Doctors photographed 10 types of data for each patient, corresponding to 10 sections: (1, 2) parasternal left ventricular long axis 2D + color, (3, 4) parasternal artery short axis 2D + color, (5, 6) apical four-chamber section 2D + color, (7, 8) apical five-chamber section 2D + color, (9, 10) subxiphoid two-chamber section 2D + color. Each section data is a video sequence, where the section is treat as a view of image in our paper. And we select five views (1,3,5,7,9) among them, resulting in 120,179 ultrasound images. We further randomly select five images (in the video) of the five views for each sample, where some views may be missing, and some views may be repeated but with different blood flow motions. Doctors also generate a corresponding diagnosis report (in English) for each patient, which belongs to either normal, VSD or ASD. Thereout, we assign the reports to the samples according to their patient id, where each report is split into different descriptions according to the pre-defined topics, like \emph{echo} and \emph{motion}. Finally, we split the image-description pairs into 19,324 training samples and 4,593 testing samples. The dataset\footnote{\url{http://mac.xmu.edu.cn/challenge/MICCAI2019-AutoGen-CDR19/index.html}} (named CardUltData) has been publicly released to promote the research.

\paragraph{\textbf{Settings.}} Common evaluation metrics for image captioning are used to provide a
quantitative comparisons. Specially, we use Bleu (B), Meteor (M), Rouge-L (R), and CIDEr (C) as \cite{chen2015microsoft}. Each of them evaluates the consistency (include co-occurrence) between the generated/candidate and real/reference descriptions. For the model architecture, we set the dimensions of all the features in RNN as 512. The sizes of the view-related and topic-related factor matrices are $5\times5$ and $10\times10$, respectively. We employ PyTorch\footnote{\footnotesize\url{http://pytorch.org}} to implement our model with maximal 60 training epochs.

\begin{center}
\makeatletter\def\@captype{table}\makeatother
\begin{minipage}{0.7\linewidth}\footnotesize\raggedleft
\caption{\footnotesize Performance comparisons of cardiovascular ultrasound report generation on CardUltData. ``B-n'' denotes the n-gram co-occurrence in Bleu metric.}
\vspace{1mm}
\begin{tabular}{c|c|c|c|c|c|c|c}
\hline
\rowcolor{mygray}
Method & B-1 & B-2 & B-3 & B-4 & C & M & R \\
\hline
CNN-RNN     & 0.859 & 0.826 & 0.800 & 0.779 & 5.966 & 0.567 & 0.850 \\ \hline
SCST-ATT    & 0.867 & 0.832 & 0.806 & 0.779 & 6.027 & 0.570 & 0.862 \\ \hline
FA-Gen      & 0.862 & 0.830 & 0.807 & 0.783 & 6.154 & 0.573 & 0.864 \\ \hline 
FE-Gen      & 0.868 & 0.837 & 0.815 & 0.786 & 6.196 & 0.580 & 0.871 \\ \hline 
FAE-Gen     & \textbf{0.873} & \textbf{0.839} & \textbf{0.820} & \textbf{0.786} & \textbf{6.217} & \textbf{0.581} & \textbf{0.877} \\ \hline
\end{tabular}
\label{tab:comparison_sota}
\end{minipage}
\makeatletter\def\@captype{figure}\makeatother
\begin{minipage}{0.28\linewidth}
\rightline{\includegraphics[width=1.0\linewidth]{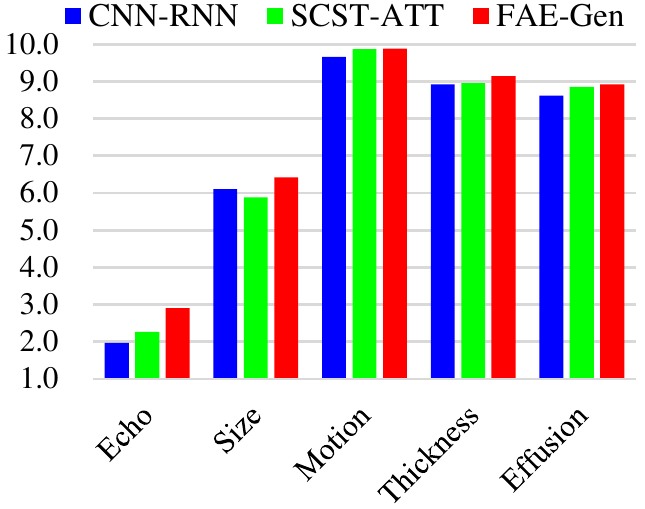}}
\setlength{\abovecaptionskip}{-3pt}
\caption{\footnotesize CIDEr evaluations on different topics.}
\setlength{\abovecaptionskip}{-4pt}
\label{fig:comparison_topics}
\end{minipage}
\end{center}
\vspace{-2mm}

\begin{figure}
\centering
\includegraphics[width=1.0\linewidth, height=0.3\linewidth]{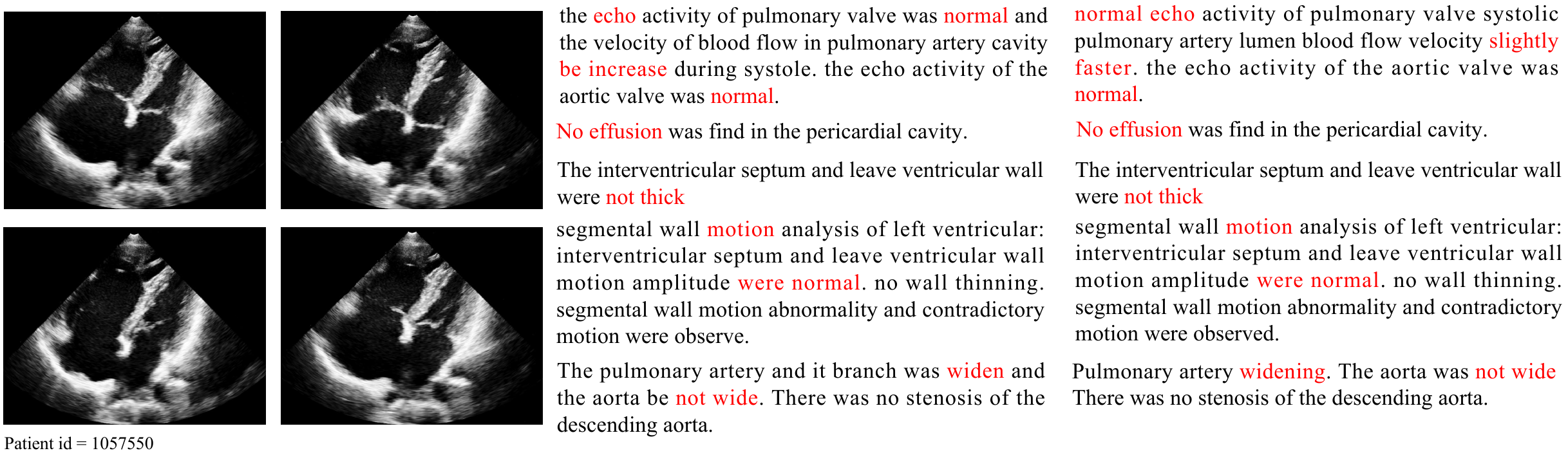}
\setlength{\abovecaptionskip}{-12pt}
\caption{Example results on CardUltData. Left: input VSD images with multiple views. Middle: the generated descriptions, Right: the ground truth (real) descriptions. Key words are marked in red. \label{fig:GeneratedDescription}}
\vspace{-4mm}
\end{figure}

\paragraph{\textbf{Evaluation and Discussion.}} We conduct the quantitative comparisons and qualitative analysis for the evaluation on cardiovascular ultrasound report generation. \textbf{First}, we compare the proposed  FAE-Gen to some respective and state-of-the-art image captioning methods, including 1) CNN-RNN \cite{vinyals2015show}, the vanilla encoder-decoder method, 2) SCST-ATT \cite{rennie2017self}, a state-of-the-art image captioning model with visual attention, 3) FE-Gen, an alternative version of FAE-Gen without view-guided factored attention, 4) FA-Gen, an alternative version of FAE-Gen without topic-oriented factored embedding. The results are shown in Tab. \ref{tab:comparison_sota}, from which we find that the proposed FAE-Gen outperforms the others over most of the metrics. Specially, FAE-Gen achieves significant improvements over FE-Genand FA-Gen, which indicates the effectiveness and superiority of both the view-guided factored attention and topic-oriented factored embedding in the unstructured-view topic-related cardiovascular ultrasound report generation. Additionally, we compare the performances on CIDEr over different topics in Fig. \ref{fig:comparison_topics}, where FAE-Gen achieve significant improvements, especially on the topic \emph{Echo}. This is probably due to the long and various syntaxes in the dataset. \textbf{Second}, we provide the the generated descriptions by FAE-Gen in Fig. \ref{fig:GeneratedDescription}, where the descriptions generated by FAE-Gen are more consistent with the ground truths, including the key pathological descriptions, like \emph{increase}, \emph{normal}, and \emph{wide}. We also find that the descriptions of some topics are consistent with the ground truth, which is probably due to the similar syntaxes of these topics in the reports.

\section{Conclusion}

In this paper, we investigate a new computer-aided medical imaging task, \emph{i.e.}, ultrasound report generation, which involves unstructured-view ultrasound images and topic-related descriptions. To this end, we propose a novel factored attention and embedding model (FAE-Gen), which mainly consists of view-guided factored attention and topic-oriented factored embedding modules. One the one hand, the view-guided factored attention module captures the homogeneous and heterogeneous morphological characteristic across different views. On the other hand, the topic-oriented factored embedding module generates the descriptions with different syntactic patterns and different emphatic contents for different topics. Experimental evaluations are conducted on a proposed large-scale clinical cardiovascular ultrasound dataset (CardUltData). Both quantitative comparisons and qualitative analysis demonstrate the effectiveness and the superiority of FAE-Gen over seven commonly-used evaluation metrics.

\newpage
\bibliographystyle{unsrt}
{\small
\bibliographystyle{splncs04}
\bibliography{mybibliography}
}

\end{document}